\documentclass[runningheads]{llncs}
\usepackage{graphicx}
\usepackage{comment}
\usepackage{amsmath,amssymb} 
\usepackage{color}
\usepackage[width=122mm,left=12mm,paperwidth=146mm,height=193mm,top=12mm,paperheight=217mm]{geometry}
\DeclareMathOperator{\E}{\mathbb{E}}
\begin{document}
\pagestyle{headings}
\mainmatter

\title{Guided Image Inpainting: Replacing an Image Region by Pulling Content from Another Image} %

\titlerunning{Guided Image Inpainting}

\authorrunning{Yinan Zhao, Brian Price, Scott Cohen, Danna Gurari}

\author{Yinan Zhao\textsuperscript{1}, Brian Price\textsuperscript{2}, Scott Cohen\textsuperscript{2}, Danna Gurari\textsuperscript{1}}

\institute{University of Texas at Austin\textsuperscript{1} \\
\email{yinanzhao@utexas.edu} \\
\email{danna.gurari@ischool.utexas.edu} \\
Adobe Research\textsuperscript{2}\\
\email{\{bprice,scohen\}@adobe.com}}

\maketitle

\begin{abstract}
Deep generative models have shown success in automatically synthesizing missing image regions using surrounding context. However, users cannot directly decide what content to synthesize with such approaches.
We propose an end-to-end network for image inpainting that uses a different image to guide the synthesis of new content to fill the hole. 
A key challenge addressed by our approach is synthesizing new content in regions where the guidance image and the context of the original image are inconsistent.  We conduct four studies that demonstrate our results yield more realistic image inpainting results over seven baselines. 
\end{abstract}

\section{Introduction} 
People often wish to replace undesired content from their photos with a plausibly realistic alternative.  For example, a visitor spends a long time trying to capture a novel pose of the Leaning Tower of Pisa but a passerby happens to occlude the tower in the captured photo. It is often desirable to remove the passerby out of the photo. 
In another case, a digital artist may wish to show Denver with a beach by replacing part of the photo with a beach from San Diego.
These tasks involve removal of undesired regions followed by automated image inpainting (also called image completion and hole filling), the task of filling in the lost part of an image. 

\begin{figure}[!ht]
\includegraphics[width=0.95\textwidth]{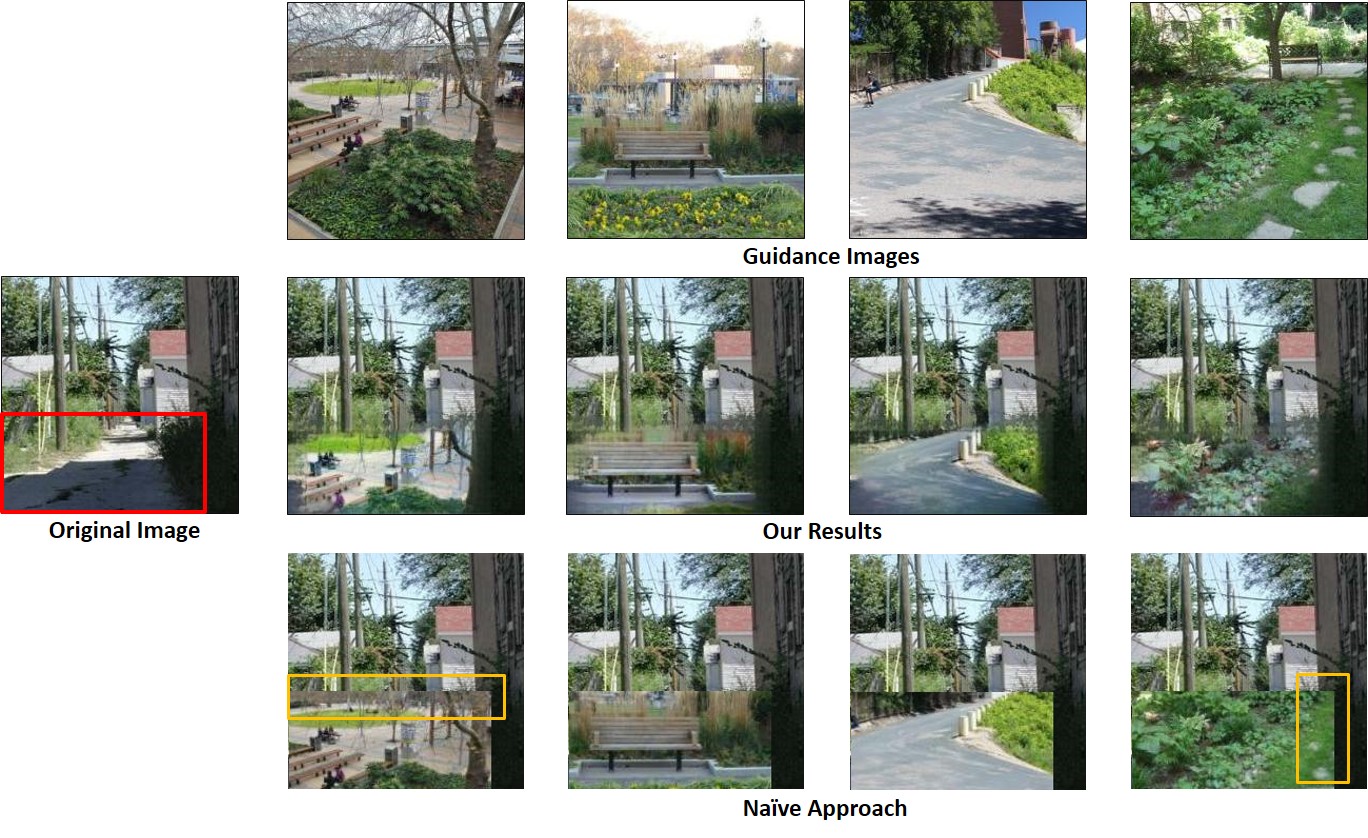}
\caption{We propose a method for automatically filling a hole in an image.  Our method takes an original image with a hole (in this example, we remove the red bounding box of the road in the original image) with a guidance image and outputs an image inpainting result.  This figure illustrates that different “guidance images” (row 1) lead to diverse image inpainting results (row 2).  We also show results from a naive approach which cuts and pastes the guidance image content into the hole (row 3).  As observed, a key challenge is synthesizing new content for the inconsistent regions in the transition between the original image and the guidance image (see yellow bounding boxes).}
\label{fig:intro_ex}
\end{figure}

Prior work has made progress on the automated image inpainting problem.  One line of work involves cutting and pasting a semantically similar patch and then blending it with the image~\cite{hays2007scene,IntRetrieval2009}.  Unfortunately, such methods typically do not generate realistic results when a pasted region does not closely match the context of the image; e.g., as exemplified in Figure \ref{fig:intro_ex} (see yellow bounding boxes on images).  Another set of methods automatically synthesize missing image regions using surrounding context~\cite{efros1999texture,efros2001image,kwatra2005texture,criminisi2003object,drori2003fragment,wilczkowiak2005hole,ctxEnc2016,yang2016high}.  With such approaches, users may receive realistic looking results, however a user cannot decide what content to synthesize in the hole.  An open challenge is how to produce a realistic looking image inpainting result while enabling the user to decide what content should fill the hole.

We introduce an image inpainting method that is designed to overcome the limitations discussed above.  Our key idea is to use another image to ``guide" the synthesis process within a deep learning framework.  The advantage of this approach is a user can directly control the result.  We illustrate this in Figure~\ref{fig:intro_ex}.  Given an original image and a region to remove, we show different ``guidance images" (top row) and results from our method (middle row).  The key difficulty in successfully using a guidance image is evolving its content to match the original image.  This difficulty is exemplified by observing results from a naive cut-and-paste method (bottom row).  The transition between the original image and pasted content often has inconsistent regions (see yellow bounding boxes on images).  Our main contribution is an end-to-end neural architecture that, given a guidance image, leverages the guidance image to fill the hole while identifying inconsistent regions and then synthesizing new content to match those regions with the image context. 

We explore three questions when evaluating our proposed inpainting approach: 1) how well does our approach capture true pixel values in the hole?, 2) how often does our approach synthesize images that people think are real? and 3) how often do people perceive our results as more realistic than alternative methods?  In our experiments, we demonstrate the effectiveness of our approach in capturing true pixel values with an image restoration task.  We also conduct human perceptual experiments that show that our approach can synthesize more realistic hole-fillings than seven alternative approaches. 
\section{Related Work}

\noindent
\textbf{Image Inpainting: }Inpainting is the process of filling in lost parts of images. Existing methods that address inpainting fall into two groups. 

The first group uses only the given image context to fill the holes. Classical inpainting \cite{bertalmio2000image,osher2005iterative} fills a hole by propagating isophotes from surrounding context to the hole via diffusion.  Texture synthesis methods \cite{efros1999texture,efros2001image,barnes2009patchmatch,kwatra2005texture,criminisi2003object,drori2003fragment,wilczkowiak2005hole} extend texture from surrounding regions to the holes. Such methods have no means to capture the semantics of the image. More recently, deep neural networks  \cite{ctxEnc2016,yeh2016semantic,iizuka2017globally,yang2016high,yu2018generative} were introduced to regress directly from surrounding context to missing pixel values. While there are multiple reasonable ways to fill the hole, these approaches do not give users the control on what should be synthesized. Zhu et al. \cite{zhu2017toward} models the distribution of possible outputs by random sampling. Our model in contrast enables the users to control the synthesis process by a guidance image. 

The second group relies on other images to fill the hole. Scene completion \cite{hays2007scene} involves cutting, pasting and blending a semantically similar patch from a database of millions of images to the original image. Internet-based inpainting \cite{IntRetrieval2009,zhu2016faithful} utilizes a more powerful web engine to improve the semantic similarity in nearest neighbor search. These approaches are good at transferring structure and high-frequency details from the guidance image to the hole, but the results often have inconsistent regions where the pasted/blended content fails to match with the context. In contrast, our approach can synthesize new consistent content in such regions.

\noindent
\textbf{Image Harmonization: }Image harmonization is the process of adjusting low-level appearances of foreground and background regions to make them compatible when generating realistic composite images. These methods \cite{sunkavalli2010multi,tao2010error,pitie2007linear,reinhard2001color,perez2003poisson,tsai2017deep} use color and tone matching techniques as well as recent learning based approaches to ensure appearance consistency. 
Blending \cite{perez2003poisson} mainly addresses color inconsistency by interpolating  the error in the transition region. Image melding \cite{darabi2012image} transforms a patch from one source to the other by patch-based synthesis. These approaches are good at transferring structure and high-frequency details from the guidance image to the hole, but in the transition region they mainly address low-level color and texture inconsistency explicitly by interpolating error or patch-based optimization. In contrast, our approach synthesizes new consistent content in the transition region using an end-to-end deep neural network. Experiments demonstrate the advantage of our system over relying on existing top-performing methods~\cite{perez2003poisson,tsai2017deep,darabi2012image} to generate image inpaintings.

\noindent
\textbf{Style Transfer: } Image style transfer \cite{gatys2016image,johnson2016perceptual,gatys2016controlling,elad2017style} has demonstrated
impressive results in example-based image stylization. It combines the high-level content of one image with the low-level style of another. Our approach is similar to style transfer in that it combines information from two images.  Our approach differs from style transfer because it inserts high-level features of the guidance patch directly into the high-level feature map of the incomplete image to synthesize  a consistent image.  Our goal is not to transfer low-level style from one image to another, but rather to synthesize a consistent hole-filling while keeping the high-level content of the guidance image. 
\section{Methods}
Our image inpainting approach is embedded in a larger system that includes three key steps.  For completeness, we provide a system overview in Section \ref{sec:systemOverview}.  Our key novelty lies in the third step, which we discuss in Sections \ref{sec:architectures} and \ref{sec:training}.  

\subsection{System Overview}
\label{sec:systemOverview}
Our system consists of three steps.  Given an image, the first step is for a user to identify a region of the image to remove.  Next, a guidance image is provided, either by a user desiring to guide the hole filling process, or by our system that automatically identifies a different image to use as guidance given the image with the user-defined region to remove.  Our implementations for these two steps are described in the experiment section (Section \ref{section5.2}).  The final component is an image inpainting approach that takes as input the \emph{incomplete image} (i.e., original image with the user-defined region removed) and the \emph{guidance image}, and returns an image inpainting result.  This final step is where the key novelty of our work lies.  We will next describe how we design a deep learning framework for guided image inpainting (Section \ref{sec:architectures}) and then discuss how we train the deep learning framework (Section \ref{sec:training}). 

\subsection{Image Inpainting Framework}
\label{sec:architectures}
\begin{figure*}[!ht]
\centering
\includegraphics[width=0.90\textwidth]{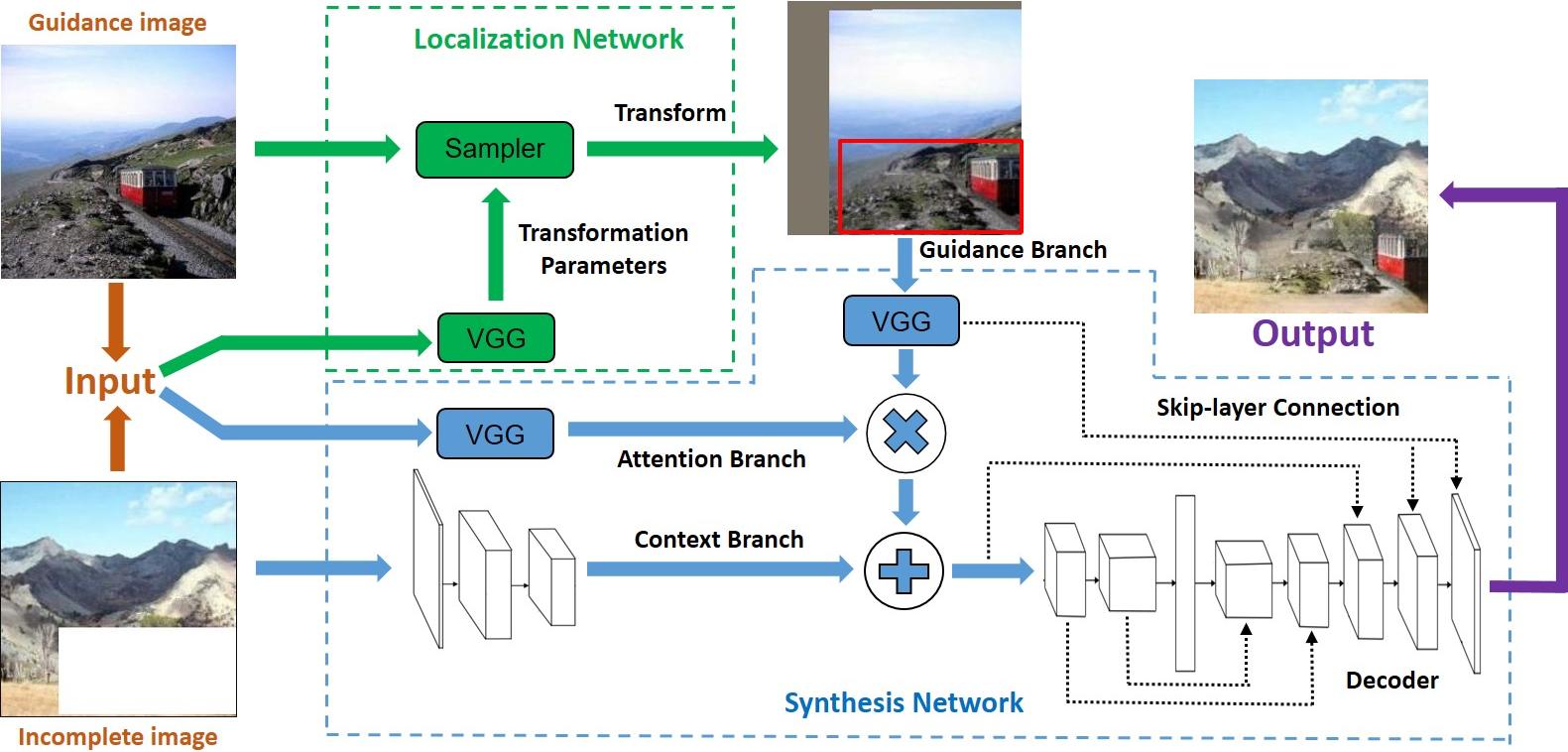}
\caption{Our proposed inpainting framework includes a 1) localization network which locates the best fitting patch in the guidance image and aligns it with the hole and 2) synthesis network which encodes both the incomplete image and the aligned guidance image to synthesize a reasonable patch to fill the hole. (Best viewed in color.)}
\label{fig:method_overall}
\end{figure*}

An overview of our deep learning framework is shown in Figure~\ref{fig:method_overall}.  Our approach takes in an \emph{incomplete image} and \emph{guidance image}, and applies two key modules sequentially to  generate an image inpainting result.  The first module identifies an image patch from the guidance image to use to replace the removed portion of the incomplete image.  We call this first step the ``Localization Network".  The second module aims to synthesize new content to fill the hole to be consistent with the image context, informed by the guidance image patch.  We call this second step the ``Synthesis Network".  We describe these two networks and how they achieve the proposed aims below.  

\subsubsection{Localization Network:}
The localization network decides how to fit a patch from the guidance image into the hole of the incomplete image using an affine transformation. We design the localization network as a spatial transformer \cite{jaderberg2015spatial} that consists of two modules. The first module uses a CNN, derived from the \emph{VGG} architecture \cite{simonyan2014very}, to regress the transformation parameters that will be used to transform the guidance image.  It takes as input an incomplete image, guidance image, and binary mask indicating the hole. All images are of size $224 \times 224$. This CNN consists of $13$ convolution layers to compute a spatial feature representation of dimensions $7 \times 7 \times 512$, followed by three fully connected layers to predict the six parameters of affine transformation. The second module is a differentiable image sampling kernel that transforms the guidance image based on the transformation parameters predicted in the first module.  After the transformation, the guidance patch that will inform the image synthesis process is aligned with the hole in the incomplete image.

\subsubsection{Synthesis Network:}
We design the synthesis network as an encoder-decoder CNN
that takes as input an incomplete image, guidance image and binary mask indicating the hole, and outputs a realistic hole filling, as illustrated in Figure \ref{fig:method_overall}. We want to take the high-level features from the aligned guidance image, directly insert them into the high level features of the incomplete image, and then decode it into a realistic hole filling. Intuitively, we expect some of the regions in the guidance image patch to be consistent with the context when inserted into the incomplete image while other regions in the guidance image will be inconsistent.  Therefore, we introduce an attention branch which computes a spatial map that indicates how well each pixel in the guidance image fits with the surrounding context in the incomplete image.  This attention branch controls to what extent to preserve the original content of the guidance image when synthesizing new content. Note that even in areas where the network is using the features from the guidance image, it still needs to adjust the color and appearance to match the surrounding context.

We design the encoder to consist of three encoding branches that are then passed to the decoder.  We use a context branch and guidance branch to encode the incomplete image and aligned guidance image separately into their high level features. We introduce an attention branch to weight the features from the aligned guidance image by multiplying the spatial attention map from the attention branch with high level features from the aligned guidance image. After that, we add together the weighted features from the guidance branch and the features from the context branch.  Next, the decoder takes the combined feature and decodes it into a realistic hole filling.  We introduce a skip-layer connection\cite{ronneberger2015u} from the guidance branch to the decoder and within the decoder with the goal to preserve details that may otherwise be lost due to the bottleneck (i.e., small number of neurons) in the encoder-decoder pipeline.  We describe below the technical details about the components in this network.

\vspace{-0.75em}
\noindent
\paragraph{Context Branch:} We use a fully convolutional network, following the \emph{VGG} architecture until \emph{conv3\_3}.  It includes six convolutional layers, resulting in a $56 \times 56 \times 256$ dimensional feature. The input to this branch is the incomplete image concatenated with a binary mask indicating the hole.

\vspace{-0.75em}
\noindent
\paragraph{Guidance Branch:} We choose a pretrained \emph{VGG} on \emph{ImageNet} until \emph{conv3\_3}.  This feature representation encodes high-level semantic features for the visual recognition task.  The feature dimension matches that of the context branch.  The input to this branch is the aligned guidance image.

\vspace{-0.75em}
\noindent
\paragraph{Attention Branch:} The attention branch has a similar architecture as the context branch, but with an additional convolutional layer followed by a sigmoid layer that outputs 1-channel feature map in the end.  This map indicates the relative weight for each spatial location.  The input is a concatenation of the incomplete image, guidance image, and binary mask indicating the hole. 

\vspace{-0.75em}
\noindent
\paragraph{Decoder:} We use as the decoder the \emph{conv3\_3} to \emph{fc7} layer from the \emph{VGG} architecture to extract a  $4096$ feature dimension.  This compact representation then goes through a series of bilinear upsampling followed by convolution to generate image completion of the original image size. Finally, we use a \emph{tanh} layer to constrain the output in the normalized range.  The input is the result from multiplying the guidance branch with the attention map and then adding that result to the context branch. 

\vspace{-0.75em}
\noindent
\paragraph{Skip-layer Connections:} We introduce a skip-layer connection~\cite{ronneberger2015u} that concatenates the \emph{conv1\_2} and \emph{conv2\_2} feature maps in the guidance branch with the corresponding symmetric feature maps in the decoder.  We also include skip-layer connections within the decoder.  These connections at different layers provide different levels of abstraction. 

\subsection{Image Inpainting Training}
\label{sec:training}
To our knowledge, there is no clear way to obtain real training data for our method.  Thus, we took inspiration from prior work that shows that models trained on synthetic data can generalize well to real images \cite{butler2012naturalistic,xu2017deep}.  We first describe a large-scale synthetic inpainting dataset we created, based on corrupting patches of the original image to use as the guidance image.  Despite the use of content from the original image and the unnatural appearance of the synthetic guidance images used in training, we will show in our experiments that our system generalizes well in real scenarios when the guidance image is a real, different image from the original image (Sections \ref{section5.2} and \ref{section5.3}).  We then describe the loss functions we use to train our localization and synthesis networks. 

\subsubsection{Training Data Generation}
\begin{figure}[!ht]
\centering
\includegraphics[width=0.95\textwidth]{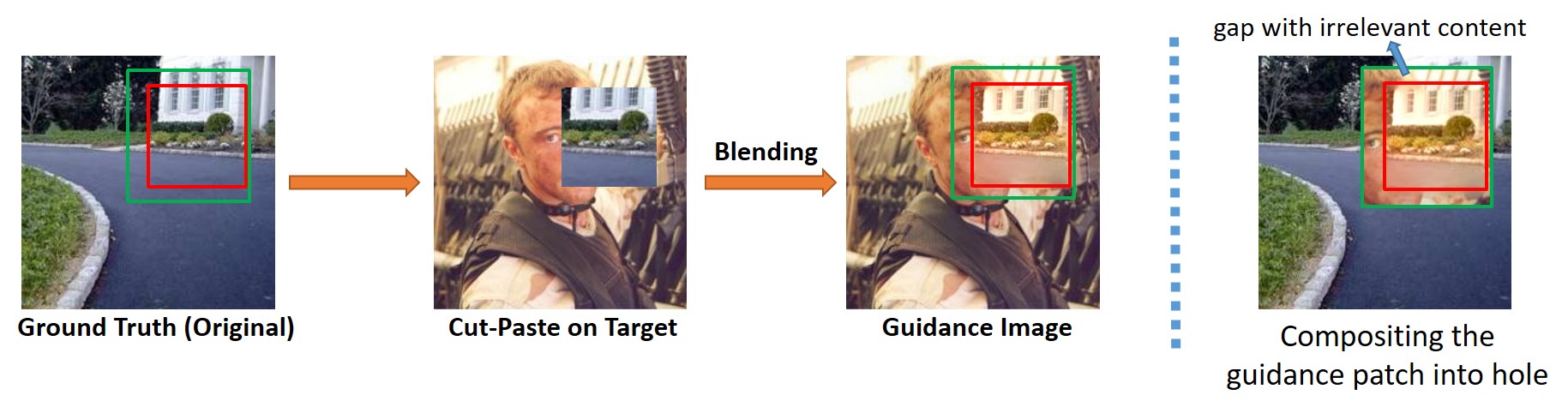}
\caption{Shown is an example illustrating our process for generating synthetic training data. After identifying the hole to remove from the original image (green bounding box) and the patch of the original image to corrupt (red bounding box), our process creates an image containing both corrupted original content (inside red bounding box) and content irrelevant to the original image (gap between red and green bounding boxes).}
\label{fig:train}
\vspace{-1em}
\end{figure}

To create a training dataset, we crop out a patch on the original image, and then corrupt the patch and use it as a guidance patch to fill the hole left on the original image, as exemplified in Figure \ref{fig:train}. The key advantage of the synthetic data is that we have ground truth inpainting result, which is the original image. Specifically, we randomly choose a patch on the original image, and then corrupt the patch by pasting and blending it into another random image (we call it a target image). Trained on the synthetic dataset in a supervised manner, the localization network is expected to localize the corrupted patch in the blended image while the synthesis network is expected to synthesize the original patch conditioned on the corrupted one.

The corruption mentioned above adjusts the color and appearance of the original patch significantly. Training on such data, our synthesis network has to learn to adjust the lighting and appearance of the guidance patch in order to synthesize a consistent hole filling.  We also make the hole on the original image (see green bounding box in Figure \ref{fig:train})  larger than the corrupted patch (see red bounding box in Figure \ref{fig:train}), and fill the gap between them with irrelevant content from the target image. The synthesis network has to learn to synthesize new content, which is not in the guidance patch, to fill that gap. It is desirable because synthesizing new content for inconsistent regions tends to make the result more realistic, and it is very likely that the inconsistent regions lie near the boundary. The gap widths are randomly chosen from a fixed range, making it possible for the synthesis network to learn where to transfer existing content in the guidance patch and where to synthesize new content.

\vspace{-1em}

\subsubsection{Image Synthesis Loss}
We describe here the loss function we use to train our synthesis network. The choice of loss function is important because the problem is underconstrained. Recent work \cite{johnson2016perceptual,dosovitskiy2016generating,ledig2016photo} has shown perceptual loss computed on feature representations of a visual perception network can be used to generate high-quality images.  Generative Adversarial
Networks (GANs) \cite{goodfellow2014generative} are shown to be successful in image generation as an adaptive loss \cite{ctxEnc2016,isola2016image,ledig2016photo}. Consequently, we combine a perceptual loss computed on a visual perception network $\phi$ together with an adversarial loss to train the synthesis network:

\vspace{-0.3em}
$$l_g(G)=\frac{1}{N}\sum_{(\tilde{I}, I_{guide}) }l_{perc}^{\phi}(I,G(\tilde{I}, I_{guide}))+\lambda_{adv} l_{adv}(G(\tilde{I}, I_{guide})) \vspace{-0.3em}$$
The number of training pairs is $N$. We use $I$ to denote the ground truth and $\tilde{I}$ for the corresponding incomplete image with the user defined region removed. $I_{guide}$ is a guidance image. The perceptual loss guides the image synthesis to match the ground truth, while the adversarial loss encourages the network to favor image generation that resides on the natural image manifold.

\paragraph{Perceptual Loss: } We compute the perceptual loss on multiple layers of a visual perception network $\phi$.  Our intention is that matching both high-level and low-level representations guides the synthesis network to learn both global structure and fine-grained details \cite{chen2017photographic}. Let $\phi_j(x)$ be the feature map of the $j$th layer of a pretrained network $\phi$ given input image $x$. $\phi_j(x)$ has shape of $C_j \times H_j \times W_j$. Suppose the completed image by our network is $\hat{I}=G(\tilde{I},I_{guide})$. The perceptual loss is defined as follows:
\vspace{-0.1em}
$$l_{perc}^{\phi}(I,\hat{I})=\sum_{j}\frac{\lambda_{j}}{C_j \times H_j \times W_j}||\phi_j(I)-\phi_j(\hat{I})||_2^2 \vspace{-0.4em}$$
The hyperparameters $\lambda_{j}$ balance the contribution of each layer $j$ to the perceptual loss. For layers $\phi_j$ we use \emph{conv1\_2}, \emph{conv2\_2}, \emph{conv3\_3}, \emph{conv4\_3}, \emph{conv5\_3} in \emph{VGG-16}. $\lambda_{j}$ are updated to normalize the
expected contribution of each layer in a certain number of iterations.

\paragraph{Adversarial Loss: }Following \cite{goodfellow2014generative} we train a discriminator network $D$ to distinguish synthesized images from natural images. We alternately optimize $D$ and a generator $G$ to solve the following min-max problem:
\vspace{-0.1em}
$$\min_{G}\max_{D}\E_{I\sim p_{data}(I)}[\log D(I)]+\E_{\hat{I}\sim p_{G}(\hat{I})}[\log (1-D(\hat{I}))] \vspace{-0.1em}$$
With this formulation, the generator $G$ is trained to fool the differentiable discriminator $D$, creating solutions highly similar to natural images. 

The discriminator $D$, following \cite{iizuka2017globally}, has two critics: a global one and a local one. The global critics takes as input the whole completed image to judge global consistency of a scene while the local critics only takes a patch centered around the completed region to encourage local consistency with the surrounding area. When training the generator, the adversarial loss $l_{adv}(\hat{I})$ is defined based on the real/fake probability of the discriminator $D$ as:
\vspace{-0.2em}
$$l_{adv}(\hat{I})=-\log D(\hat{I})$$

\subsubsection{Localization Loss}
We now describe the loss function we use to train our localization network. Although the spatial transformer has a differentiable sampling kernel, the function of image synthesis with respect to the transformation parameters does not have a good shape that is easy to be optimized. To make the network trainable, we add a loss directly on the predicted transformation parameters (we call it localization loss). 

Suppose we sample $M$ points in the guidance image, and $x_i(I)$ is a vector representing  homogeneous coordinate of the $i$-th sampled point. $T(I)$ is the ground truth affine transformation matrix for image $I$, which is known when we create our dataset. $\bar{T}(I)$ is the estimated transformation matrix. We define the localization loss $l_{loc}$ as the average of Euclidean distances between estimated point positions after transformation and their corresponding ground truth locations:
\vspace{-0.4em}
$$l_{loc}(\bar{T})=\frac{1}{MN}\sum_{I}\sum_{i=1}^{M}||\bar{T}(I)x_i(I)-T(I)x_i(I)||^2_2 \vspace{-0.4em}$$
We encourage the estimated transformation matrix to be close to the ground truth transformation by minimizing the localization loss.

\subsubsection{Implementation}
We create the synthetic dataset from ADE20K \cite{zhou2017scene}. There are $20,210$ images in the training set of ADE20K and $2000$ in the validation set. For each image in the training set, we generate 50 pairs of incomplete image and guidance image, resulting in $1,010,500$ pairs for training. The localization network and synthesis network are trained separately with localization loss and synthesis loss. Training them jointly is undesirable since this leads the synthesis network to blur the image generation to accommodate localization errors.

For training CNNs, we use Tensorflow \cite{abadi2016tensorflow} framework. We choose Adam \cite{kingma2014adam} as the optimization method. We fix the parameters of Adam as $\beta_1=0.5$ and $\beta_2=0.999$. The initial learning rate in Adam is set to $l_r=0.0002$.  

We use a weighted sum of perceptual loss and adversarial loss to train the generator. The weight of adversarial loss is $\lambda_{adv}=2$. $M=8$ points are sampled in each image to compute the localization loss. We train the synthesis network for $1,918,000$ iterations with batch size equal to 2. It takes roughly two weeks to train on NVIDIA GeForce GTX 1080 GPU. The localization network is trained for $841,000$ iterations, which takes an extra 5 days.

\section{Evaluation}
We now describe our studies to evaluate the power of our localization and synthesis network in localizing fitting patches and synthesizing realistic images. We address the following research questions:
\vspace{-0.30em}
\begin{itemize}
  \item How well does our model capture true pixel values in the missing region?
  \item How often does our method synthesize realistic images that humans cannot distinguish from natural images?
  \item How often is our synthesis preferred to baselines?
\end{itemize}
We conduct four experiments to address these questions, evaluating synthesis and localization separately. In Section~\ref{section5.1}, we measure how well our synthesis network restores true pixel values of the original patch given a corrupted one as guidance. Section~\ref{section5.2} demonstrates the effectiveness of our synthesis network to synthesize images that people mistakenly perceive as real. In Section~\ref{section5.3}, we conduct pairwise comparisons between our synthesis and baselines to measure how often our results are preferred to alternative methods. In Section~\ref{section5.4}, we perform pairwise comparisons between our localization and Local Context Matching \cite{hays2007scene} to investigate which localization method results in more realistic synthesis.

\subsection{Image Restoration}\label{section5.1}
\noindent
\textbf{Baselines: }We compare our synthesis network to seven related baselines across five types of methods.  We use the code provided by the authors.

\begin{itemize}
\item \textit{Texture synthesis}: Content-Aware-Fill (CAF) using PatchMatch \cite{barnes2009patchmatch}.
\item \textit{Deep Generative Models}: We evaluate three methods for this type of approach.  We use Context Encoder (CE) \cite{ctxEnc2016}, where we retrain it with \emph{l2} and adversarial loss on ADE20K \cite{zhou2017scene}
.  We also use High-Resolution Image Inpainting (HR) \cite{yang2016high} and Globally and Locally Consistent Image Completion (GLCIC) \cite{iizuka2017globally}.  
\item \textit{Harmonization}: Deep Harmonization (DH) \cite{tsai2017deep}.
\item \textit{Blending}: Poisson Blending (PB) \cite{perez2003poisson}.
\item \textit{Melding}: Image Melding (IM) \cite{darabi2012image}.
\end{itemize}

CAF, CE, HR and GLCIC hallucinate the missing region from the context without guidance. DH, PB, IM and our method use a corrupted patch from the original image as guidance. The latter four approaches have access to additional guidance images that are not available to the former four approaches. We compare to both classes of algorithm for a comprehensive evaluation. 

\noindent
\textbf{Dataset: }We evaluate on our synthetic test set, consisting of 2000 images with random holes derived from ADE20K validation set. The guidance patch is corrupted from the original patch, with a gap filled with unrelated content near the boundary. 
To evaluate the synthesis network exclusively, we assume ground truth localization, aligning the guidance patch with the hole perfectly.

\noindent
\textbf{Evaluation Method: }We evaluate the restoration by measuring pixelwise distance between the prediction and ground truth using three metrics: L1 Loss, L2 Loss and PSNR. 

\noindent
\textbf{Results: }  Results are shown in Table \ref{table:PSNR}.  As observed, our synthesis network consistently offers significant gains over baselines for all three metrics. Our synthesis method decreases the error by $\textbf{5.34\%}$ and $\textbf{1.93\%}$ in terms of Mean L1 Loss and Mean L2 Loss compared to the best baseline, achieving $\textbf{6.89\%}$ and $\textbf{1.11\%}$ respectively. We also outperform the best baseline by $\textbf{4.13dB}$ in PSNR, achieving $\textbf{20.68dB}$.  Interestingly, our results show that methods that hallucinate results without guidance (CE, HR, GLCIC) perform similarly to those that have guidance (PB, IM).  This highlights both the effectiveness of the methods and the difficulty in properly leveraging the information in a corrupted patch.  The results reveal our method is able to effectively utilize this corrupted information.
We attribute the success in this task to the nature of our approach: it can transfer the structure of the guidance image with color and appearance adjusted and synthesize appropriate new content when necessary (fill the gap with irrelevant content in this case).

\begin{table}[!ht]
\label{table:PSNR}
\begin{center}
\begin{tabular}{c c c c |  c c c c}
\hline
Method & Mean L1 & Mean L2 & PSNR &  Method & Mean L1 & Mean L2 & PSNR\\
\hline\hline
CAF\cite{barnes2009patchmatch} & 15.43\% & 5.09\% & 14.38dB & PB\cite{perez2003poisson} & 13.63\% & 3.28\% & 15.41dB \\
CE\cite{ctxEnc2016} & 12.91\% & 3.21\% & 15.91dB & IM\cite{darabi2012image} & 12.23\% & 3.04\% &  16.55dB\\
HR\cite{yang2016high} & 13.05\% & 3.29\% & 15.83dB & GLCIC\cite{iizuka2017globally} & 13.28\% & 3.47\% & 15.56dB \\
DH\cite{tsai2017deep}  & 18.87\% & 6.02\% & 12.73dB & Ours & \textbf{6.89\%} & \textbf{1.11\%} & \textbf{20.68dB} \\
\hline
\end{tabular}
\end{center}
\caption{Results of our method and seven baselines with respect to three metrics for the image restoration task.  Smaller Mean L1 values, smaller Mean L2, and larger PSNR values indicate better performance. (\% in the table is included to facilitate reading).
}
\label{table:PSNR}
\end{table}

\subsection{Absolute Realism}\label{section5.2}
In this section, we show people a random image from a mixed dataset of real and synthesized images and ask them to guess whether the shown image is real or fake.  We balance the number of real and fake images to avoid a strong real/fake prior.  We investigate using a guidance image that is a different real image from the original image to address the common use case scenario of replacing content in an image with content from another image.

To recruit people for our human perception task, we employ crowd workers on Amazon Mechanical Turk (AMT).  We show each image for four seconds to give people sufficient time to determine its realism.  Our choice of this time period is motivated in part by the work of Joubert et al. \cite{joubert2007processing}, which showed that humans can understand scenes and complete simple tasks such as categorizing “man-made” and “natural” scenes with high accuracy (both around 96\%) in around 390ms.
\noindent{\textbf{Implementation:}}
We implement the components in Section \ref{sec:systemOverview} as follows.  

We use a rectangular hole to remove content from the original image, since bounding boxes are an efficient, commonly-used approach to annotate images.  We simulate a user who wants to remove an object or part of a scene by using a ground truth semantic segmentation to select the region to remove and fitting a bounding box around it.

For a guidance image, we are motivated to efficiently find an image that is semantically similar to the original image.  We follow the work of Heo et al.~\cite{heo2016shortlist} to retrieve nearest neighbors in a large dataset.  We examine two retrieval scenarios: using (a) the full original image (b) or the incomplete image with its hole filled by \textit{Content-Aware-Fill} \cite{barnes2009patchmatch}. We will report results for Retrieval (a) and (b) separately. In Retrieval (a), we assume access to the full image because the hole often comes from replacing part of the original image. 
In Retrieval (b) we remove the original content in the hole to prevent it from biasing the retrieval. 

\noindent
\textbf{Baselines: }We use the same seven baselines as in Section \ref{section5.1}. 

\noindent
\textbf{Dataset:} We use images in ADE20K validation set as original images and retrieve guidance images in MSCOCO~\cite{lin2014microsoft}. To evaluate synthesis separately, we use Local Context Matching (LCM)~\cite{hays2007scene} as localization for all the synthesis methods. LCM~\cite{hays2007scene} minimizes pixelwise SSD error in the local context. We randomly sample 100 images to go through our synthesis network and all the baselines. Inpainting results from all the methods are mixed together with 300 real images from ADE20K validation set to balance the number of real and fake images. 

\noindent
\textbf{Evaluation Method: }We measure absolute realism of the synthesized images by the fraction of images deemed to be real in the AMT study.

\noindent
\textbf{Results: }As observed in Table \ref{table:real_fake}, although it is still far from natural images, $\textbf{33\%}$ and $\textbf{36\%}$ of the synthesized images by our approach are deemed to be real with Retrieval (a) and (b) respectively. It outperforms all the baselines significantly except for only slightly beating \textit{Content-Aware-Fill(CAF)} in Retrieval (a). We will show in Section \ref{section5.3} that our synthesized images are rated to be more realistic than \textit{Content-Aware-Fill} in a more careful pairwise comparison. Despite our synthesis network being trained on a synthetic dataset, our results show it can generalize well to real images.

\begin{table}[!ht]
\begin{center}
\begin{tabular}{| c | c | c | c | c | c | c | c | c | c |}
\hline
Method & NI & CE\cite{ctxEnc2016} & HR\cite{yang2016high} & CAF\cite{barnes2009patchmatch} & PB\cite{perez2003poisson} & DH\cite{tsai2017deep} & GLCIC\cite{iizuka2017globally} & IM\cite{darabi2012image} & Ours\\
\hline
Retrieval (a) & 97.7\% & 10.0\% & 14.0\% & 31.0\% & 18.0\% & 23.0\% & 14.0\% & 23.0\% & \textbf{33.0\%}\\ \hline
Retrieval (b) & 97.7\% & 22.0\% & 15.0\% & 16.0\% & 20.0\% & 22.0\% & 12.0\% & 27.0\% & \textbf{36.0\%}\\
\hline
\end{tabular}
\end{center}
\caption{User study results showing the perceived absolute realism of synthesized images. The numbers in the table are the percentage of images deemed to be real. NI refers to natural images. Retrieval (a) uses the full original image to retrieve the guidance image while Retrieval (b) uses the incomplete image with its hole filled by CAF \cite{barnes2009patchmatch} to retrive the guidance image from a large database.}
\label{table:real_fake}
\end{table}

\vspace{-1em}
\subsection{Relative Realism}\label{section5.3}
We next investigate how often the generated images by our synthesis network are considered to be more realistic than those created by baselines. We conduct pairwise comparisons between our method and all the baselines in the aforementioned absolute realism study. Each pair contains two images synthesized by our synthesis network and a baseline for the same incomplete image using the same guidance image. The users are asked to select the more realistic image after they are shown a pair side-by-side for an unlimited time. 

\noindent
\textbf{Implementation: } We use the same implementation as in Section \ref{section5.2}. 

\noindent
\textbf{Baselines: } We use the same seven baselines as in Section \ref{section5.2}. 

\noindent
\textbf{Dataset: } We use the same dataset as in Section \ref{section5.2}.

\noindent
\textbf{Evaluation Method: }We compute the relative realism by the percentage of pairs in which our synthesized images are considered to be more realistic by people (i.e., AMT crowd workers) than the baselines.

\noindent
\textbf{Results: }As shown in Table \ref{table:relative_score}, people rate our synthesized images more realistic than baselines for at least $\textbf{66\%}$ of test images. For example, our results are more realistic than \textit{Content-Aware-Fill}, which is the best baseline in the absolute realism experiment for $\textbf{70\%}$ of images (using Retrieval method (a) to find a guidance image). These results reveal greater improvements of our synthesis over baselines than is evident from the absolute realism study. Our results are similar whether using Retrieval method (a) or (b) to choose a guidance image. 

\begin{figure*}[!ht]
\includegraphics[width=0.985\textwidth]{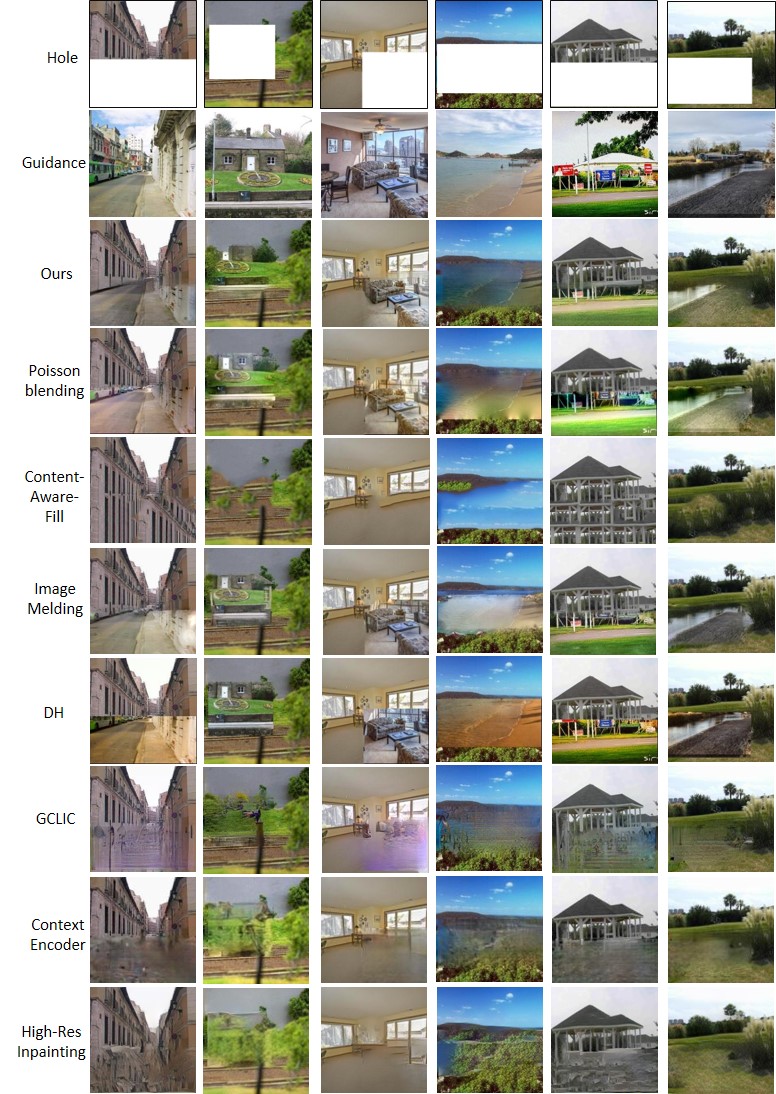}
\caption{Visual comparison of inpainting results by different methods. Compared to  methods that do not use guidance images (\textit{CAF} \cite{barnes2009patchmatch}, \textit{CE} \cite{ctxEnc2016}, \textit{GLCIC} \cite{iizuka2017globally} and \textit{HR} \cite{yang2016high}), our results look more realistic by transferring matching content of the guidance image to the hole. Compared to the baselines that use guidance images (\textit{PB} \cite{perez2003poisson}, \textit{DH} \cite{tsai2017deep} and \textit{IM} \cite{darabi2012image}),  our approach is better at avoiding artifacts along the boundaries of the hole by synthesizing new content for inconsistent regions.}
\label{fig:results}
\end{figure*}

We show qualitative results from our results and the baselines in Figure \ref{fig:results}. As observed, our synthesis network generates more realistic hole-fillings that transfer both high-level structure and necessary fine-grained details from the guidance image, while maintaining consistency with the context of the original image. We will show more qualitative comparisons in the appendix.

Our approach can also synthesize diverse inpainting results given different guidance images. We will demonstrate the diversity with qualitative results in the appendix.

\begin{table}[!ht]
\begin{center}
\begin{tabular}{| c | c | c | c | c | c | c | c |}
\hline
Method & HR\cite{yang2016high} & PB\cite{perez2003poisson} & DH\cite{tsai2017deep} & CAF\cite{barnes2009patchmatch} & CE\cite{ctxEnc2016} & IM\cite{darabi2012image} & GLCIC\cite{iizuka2017globally}\\
\hline
Retrieval (a) & 76\% & 76\% & 71\% & 70\% & 70\% & 67\% & 66\%\\ \hline
Retrieval (b) & 71\% & 73\% & 72\% & 70\% & 73\% & 67\% & 70\%\\
\hline
\end{tabular}
\end{center}
\caption{Shown are results from our relative realism study. Each cell lists the percentage of pairs in which synthesized images by our approach are rated more realistic than the corresponding baseline in human perception experiments. Chance is at $50\%$ result.}
\label{table:relative_score}
\end{table}

\subsection{Localization}\label{section5.4}
We evaluate the synthesis network in the previous three sections. In this section, we evaluate our localization method by an ablation study. We use our synthesis network to fill in holes for all localization methods. To draw a direct comparison between our method and baselines, we conduct a relative realism user study. Each pair that is compared contains inpainting results for the same incomplete image and guidance image, but using different localization methods. People are asked to choose the more realistic image after the two images are shown side-by-side for an unlimited time. 

\noindent
\textbf{Baseline: }We compare to \textit{Local Context Matching} \cite{hays2007scene}, which minimizes pixelwise SSD error in \textit{Lab} color space. It has no knowledge of global structure or semantics. 

\noindent
\textbf{Dataset:} We use images in the validation set of ADE20K as incomplete images and use the full images to retrieve semantically similar images as guidance from MSCOCO.  We randomly sample 50 images in ADE20K for evaluation, each with 10 retrieved neighbors in MSCOCO as guidance images. Thus, each hole has 10 different hole-fillings. It is more robust to retrieve multiple guidance images than a single one to evaluate localization because it is possible that the guidance image does not contain a good matching patch for hole filling. There are 500 relative comparisons. 

\noindent
\textbf{Evaluation Method: }We use the same approach as in Section \ref{section5.3}.

\noindent
\textbf{Results: }
The synthesized images based on our localization is rated to be more realistic than \textit{Local Context Matching} in $\textbf{53.8\%}$ of all the pairwise comparisons. We conjecture that the gain comes from consideration of global structure. Our localization network processes the whole image while \textit{Local Context Matching} only takes into account local information in the context. The gain is relatively small probably because the distribution of features for localization in the synthetic images is different from that in natural images. The localization network trained on the synthetic dataset does not generalize as well to natural images.
\section{Conclusion}
We introduce an end-to-end inpainting model to localize a matching patch in the guidance image and transfer its content to the hole followed by synthesizing a realistic hole-filling. Despite training on a synthetic dataset, our synthesis network generalizes well to natural images. The human perceptual experiments show that our approach synthesizes more realistic images than all the baselines.

\vspace{1em}
\noindent
\textbf{Acknowledgements} We thank the anonymous crowd workers for participating in our experiments.

\clearpage

\bibliographystyle{splncs}
\bibliography{egbib}
\section{Appendix}
\subsection{Image Restoration}

\begin{figure}
\centering
\includegraphics[width=0.98\textwidth]{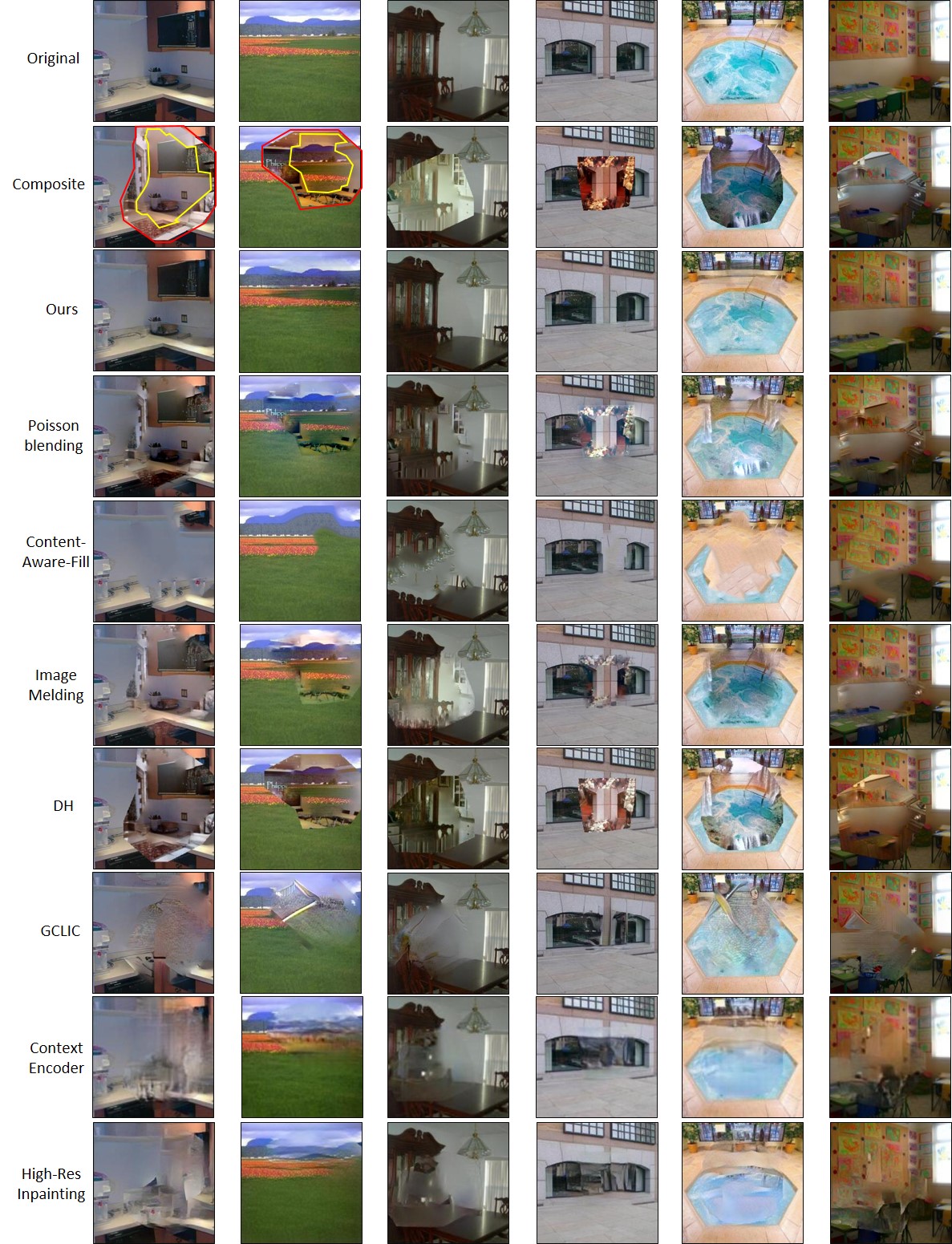}
\caption{Shown are image restoration results for our approach and seven baselines \cite{ctxEnc2016,perez2003poisson,tsai2017deep,darabi2012image,yang2016high,iizuka2017globally,barnes2009patchmatch} on the synthetic test set with holes of arbitrary shape. We create the synthetic dataset by corrupting patches of the original image to use as guidance patches. The composite images (row 2) are obtained by pasting the corrupted guidance patches onto the holes directly. We fill the gap between the corrupted patch and the hole with irrelevant content. We show the corrupted patch (yellow region) and the hole (red region) for the first two examples (column 1 and 2). Our method can adjust the color and appearance of consistent regions and synthesize new content to fill in the inconsistent gap (the region between red and yellow) near the boundary.}
\label{fig:rand_val}
\end{figure}

\begin{figure}
\centering
\includegraphics[width=0.98\textwidth]{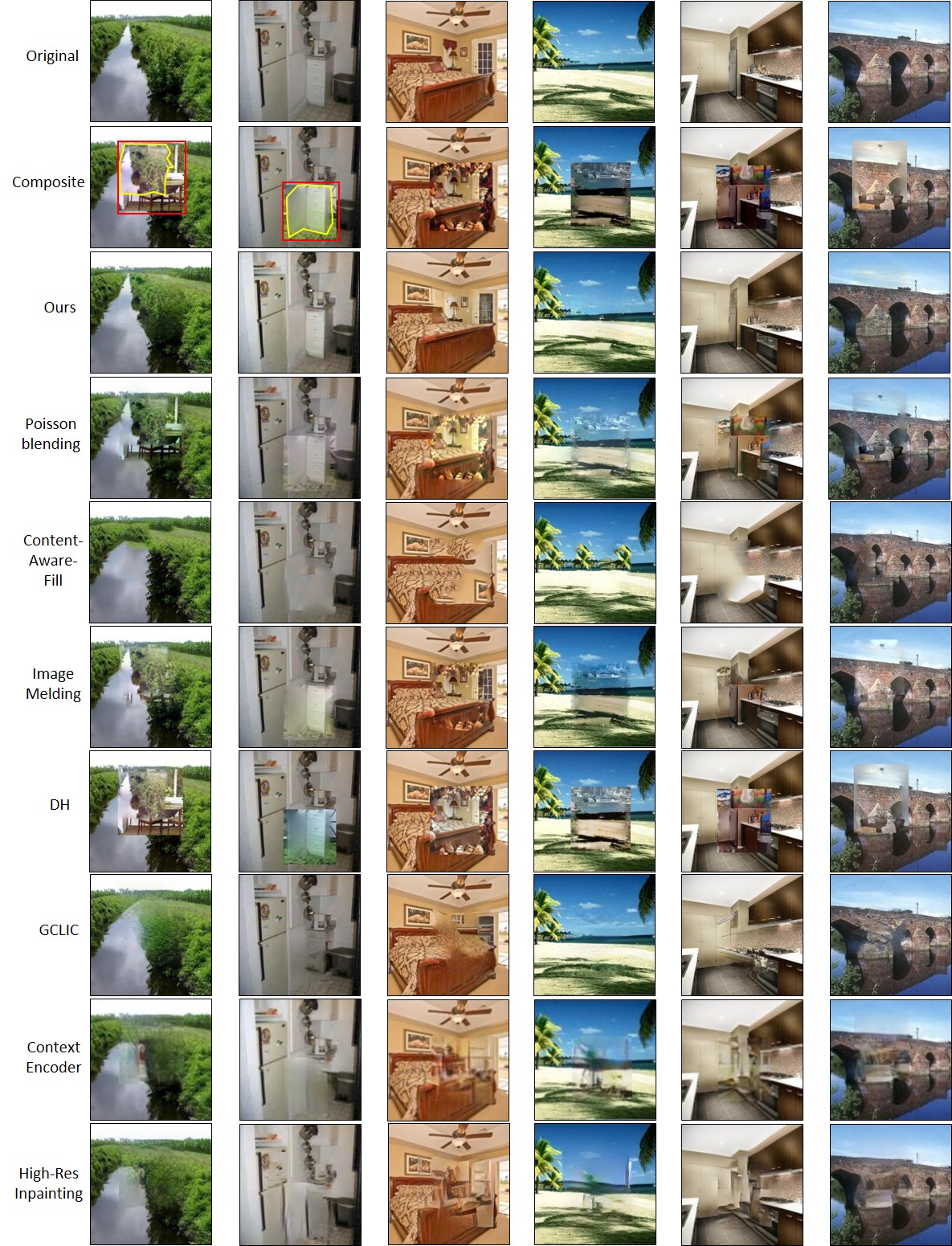}
\caption{Shown are image restoration results for our approach and seven baselines \cite{ctxEnc2016,perez2003poisson,tsai2017deep,darabi2012image,yang2016high,iizuka2017globally,barnes2009patchmatch} on the synthetic test set with bounding box holes. We create the synthetic dataset by corrupting patches of the original image to use as guidance patches. The composite images (row 2) are obtained by pasting the corrupted guidance patches onto the holes directly. We fill the gap between the corrupted patch and the hole with irrelevant content. We show the corrupted patch (yellow region) and the hole (red region) for the first two examples (column 1 and 2). Our method can adjust the color and appearance of consistent regions and synthesize new content to fill in the inconsistent gap (the region between red and yellow) near the boundary.}
\label{fig:rand_val1}
\end{figure}

In Section \ref{section5.1}, we compare our method to seven baselines quantitatively in the task of image restoration. We show visual results in Figure \ref{fig:rand_val} and Figure \ref{fig:rand_val1} to exemplify our performance in the image restoration task with holes of arbitrary shape and bounding box.  

As observed, our method can adjust the color and appearance of consistent regions and synthesize new content to fill in the inconsistent gap near the boundary. For example, in column 1, row 2 of Figure \ref{fig:rand_val}, the window and floor texture (the region between red and yellow) are inconsistent with the context of the original image. In column 1, row 3 our result shows that our approach can synthesize the wall and countertop textures so those regions match the context. 

Similarly in Figure \ref{fig:rand_val}, our result in column 2, row 3 shows that our approach synthesizes grass texture for the inconsistent region (the region between red and yellow). Note that the shape and width of the inconsistent gap are random in our training dataset. Therefore, the shape and width of the inconsistent regions (between red and yellow) in column 1 and 2 are distinct and different from the other examples shown in columns 3-6. Our model has to learn to identify inconsistent regions and synthesize new content for those regions.

\begin{figure}
\centering
\includegraphics[width=1.0\textwidth]{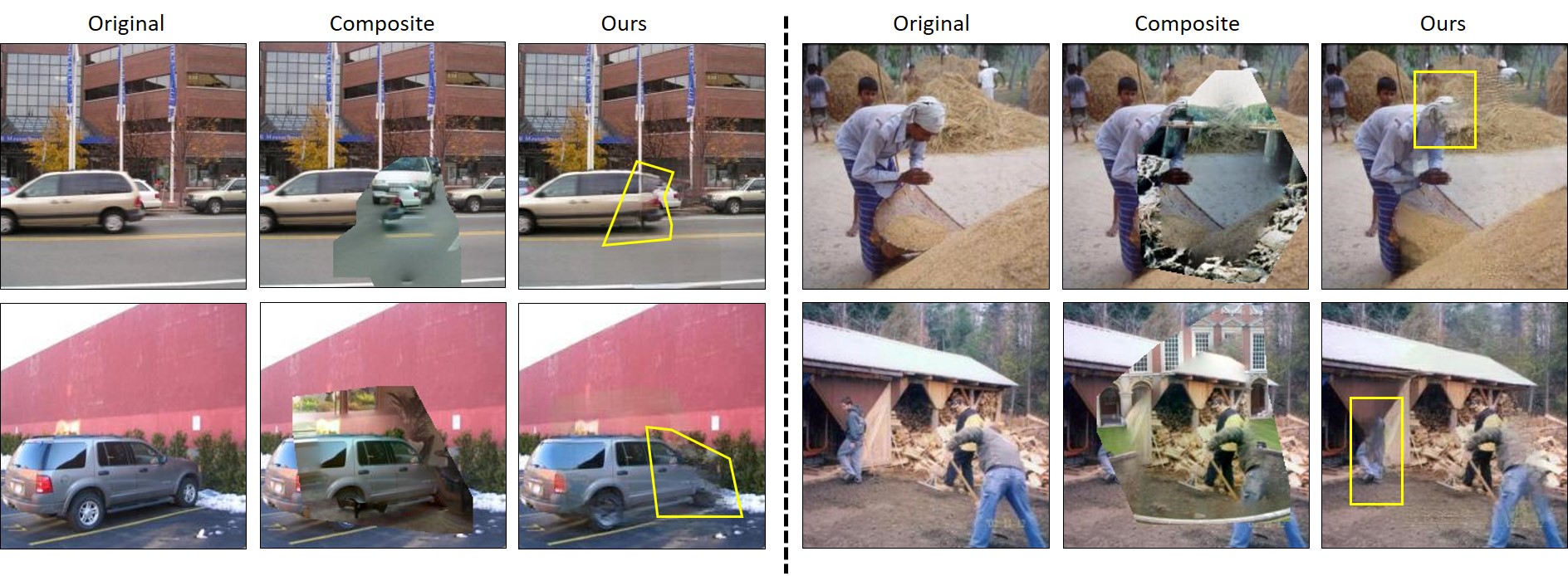}
\caption{We show failure cases of our synthesis network on image restoration. The composite images are obtained by pasting the corrupted guidance patches onto the holes directly. It is not easy for our synthesis network to synthesize complex structures such as a car and people from scratch in the inconsistent regions. Left: failure cases on cars. Right: failure cases on people. Yellow regions indicate where our synthesis fails.}
\label{fig:limitation1}
\end{figure}

We also show some failure cases in Figure~\ref{fig:limitation1}. Although our synthesis network can synthesize realistic hole-fills by transferring matching content and synthesizing new content for inconsistent regions, it is not easy for our network to synthesize complex structures from scratch in the inconsistent regions, for example car (on the left) and people (on the right) as exemplified in Figure \ref{fig:limitation1}. 

\subsection{Relative Realism}
\begin{figure}
\centering
\includegraphics[width=1.0\textwidth]{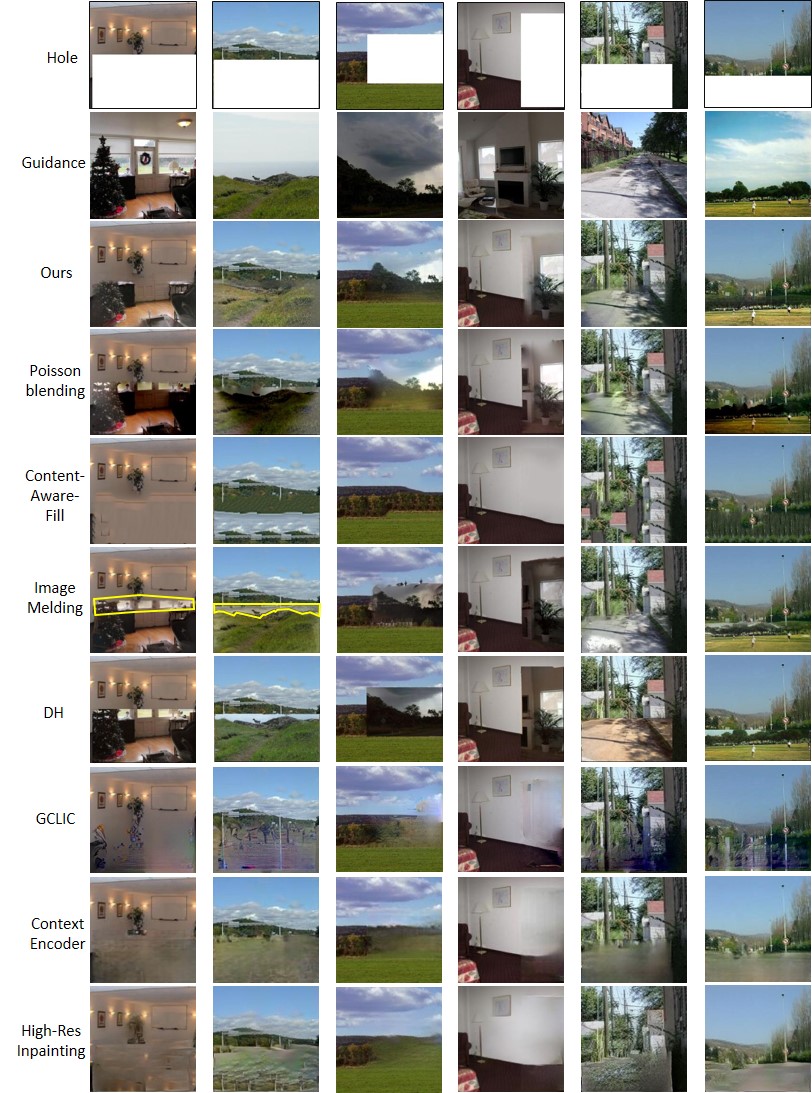}
\caption{We show a visual comparison of inpainting results by different methods. Compared to  methods that do not use guidance images (\textit{CAF} \cite{barnes2009patchmatch}, \textit{CE} \cite{ctxEnc2016}, \textit{GLCIC} \cite{iizuka2017globally} and \textit{HR} \cite{yang2016high}), our results look more realistic by transferring matching content of the guidance image to the hole. Compared to the baselines that use guidance images (\textit{PB} \cite{perez2003poisson}, \textit{DH} \cite{tsai2017deep} and \textit{IM} \cite{darabi2012image}),  our approach is better at avoiding artifacts along the boundaries of the hole by synthesizing new content for inconsistent regions (exemplified by yellow in row 6).}
\label{fig:result}
\end{figure}

In Section \ref{section5.3} Figure \ref{fig:results}, we compare our method to seven baselines~\cite{ctxEnc2016,perez2003poisson,tsai2017deep,darabi2012image,yang2016high,iizuka2017globally,barnes2009patchmatch} in terms of relative realism. We show more examples to qualitatively exemplify the comparison in Figure \ref{fig:result}. Note that we use \textit{Local Context Matching}~\cite{hays2007scene} as localization for all the synthesis methods in order to evaluate synthesis separately. 

Our approach can synthesize more realistic results because it uses guidance information to help fill the hole wherever the content is consistent and synthesizes new content where it is not consistent. In Figure \ref{fig:result}, the window region in column 1, row 6 and sky region in column 2, row 6 (the region denoted by yellow) are inconsistent with the context of the original image. Our results in columns 1-2, row 3 show that our approach can synthesize the wall and tree textures to match the context. As observed in columns 1-2, rows 4, 6, and 7, other approaches~\cite{perez2003poisson,darabi2012image,tsai2017deep} using the guidance image leave those inconsistent regions of window and sky texture in the results.

\begin{figure}
\centering
\includegraphics[width=1.0\textwidth]{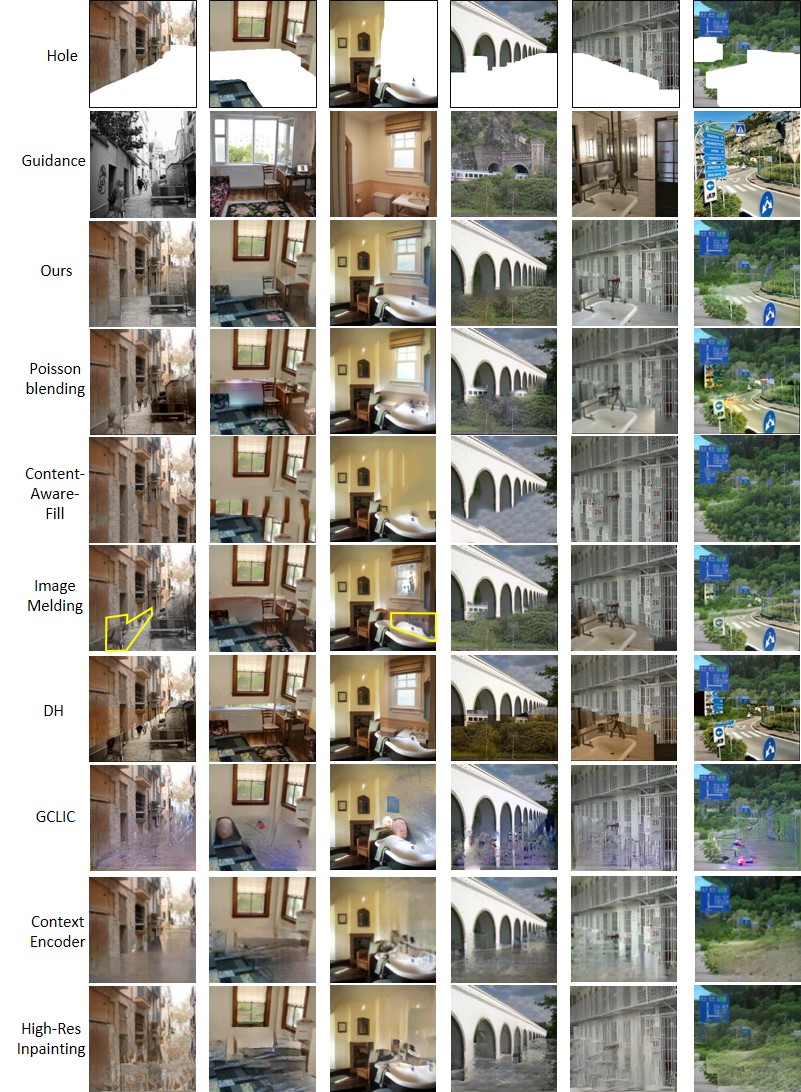}
\caption{We show a visual comparison of inpainting results by different methods on holes of arbitrary shape. We use the ground truth semantic segmentation to select part of a scene to remove. Although an arbitrary shape is less user-friendly than a bounding box, our approach can synthesize realistic hole-fills by transferring matching content of the guidance image and synthesizing new content for inconsistent regions (exemplified by yellow in row 6).}
\label{fig:result1}
\end{figure}

We also show qualitative results from our method and seven other baselines~\cite{ctxEnc2016,perez2003poisson,tsai2017deep,darabi2012image,yang2016high,iizuka2017globally,barnes2009patchmatch} on holes of arbitrary shape in Figure \ref{fig:result1}. We use the ground truth semantic segmentation to select part of a scene to remove. Although arbitrary shape is not as user-friendly as bounding box, our approach can still synthesize realistic hole-fills that are consistent with the context on holes of arbitrary shape. 

\subsection{Diverse Hole-Fills}
\begin{figure}
\centering
\includegraphics[width=0.9\textwidth]{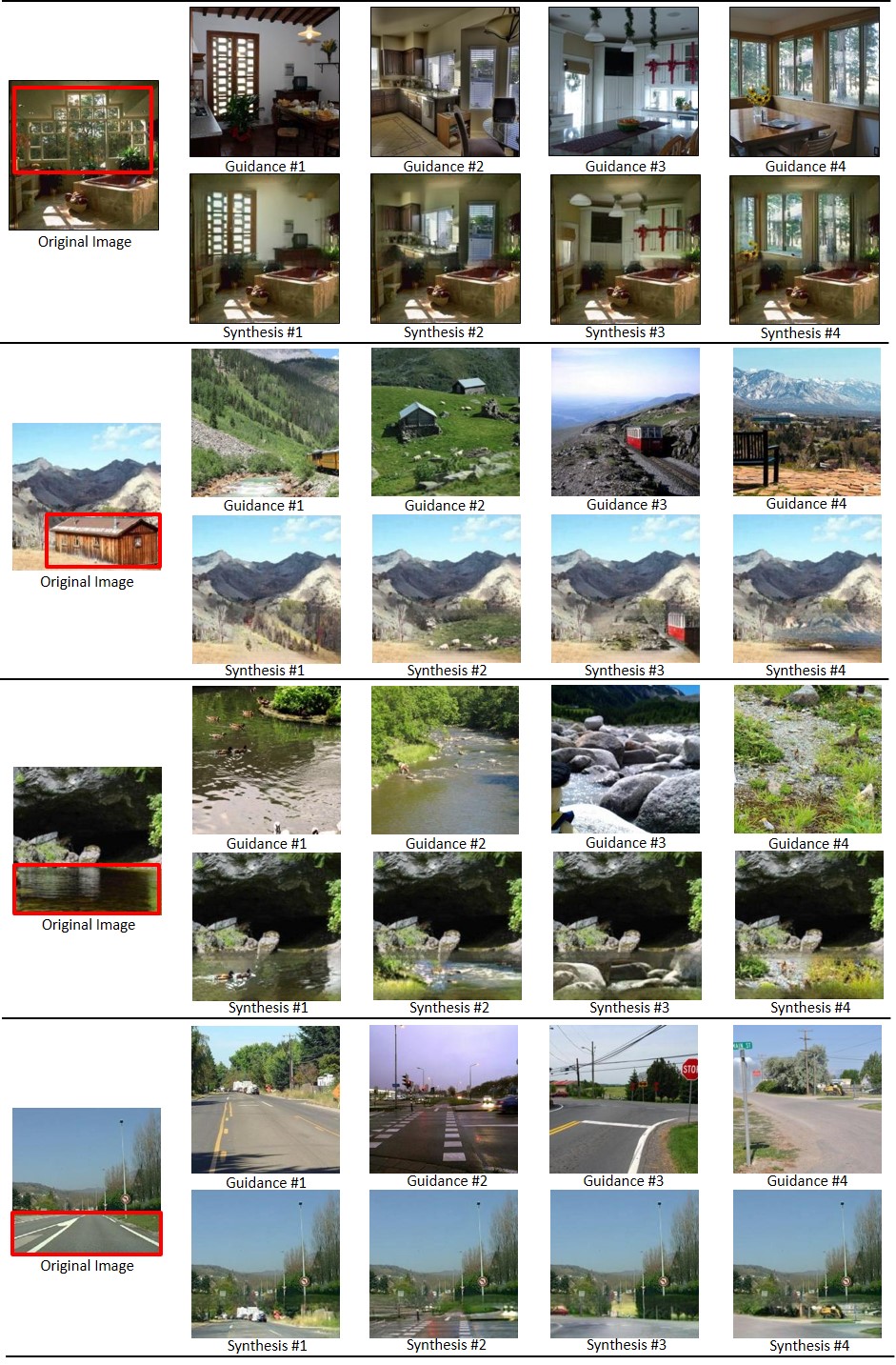}
\caption{Shown are diverse inpainting results for a hole guided by different guidance images using our inpainting architecture.  The hole is created by removing the red bounding box region in the original image. Given a guidance image, our approach first localizes a matching patch, and then synthesizes a consistent hole-filling informed by the guidance patch. }
\label{fig:multiple1}
\end{figure}

\begin{figure}
\centering
\includegraphics[width=0.9\textwidth]{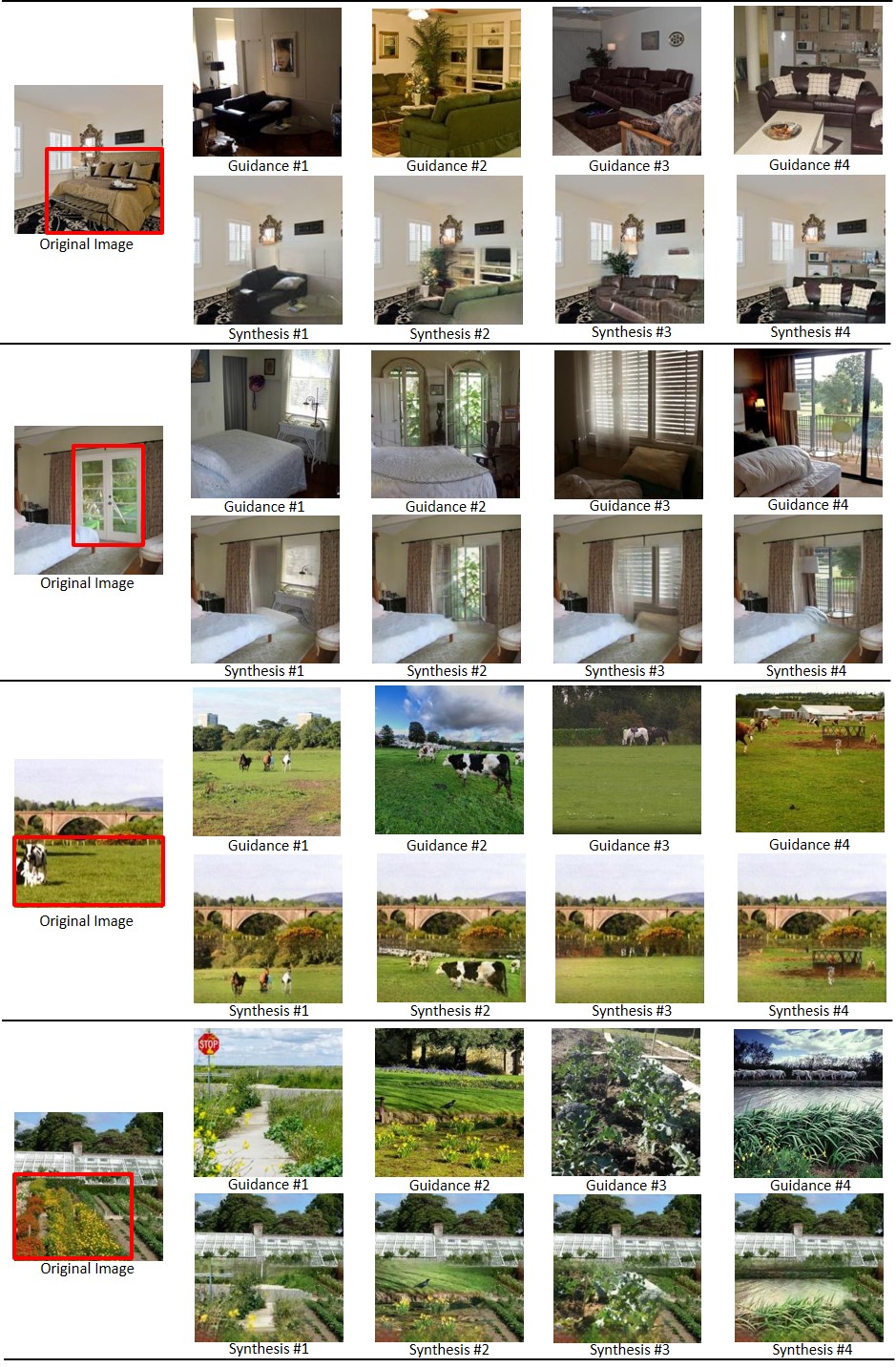}
\caption{Shown are diverse inpainting results for a hole guided by different guidance images using our inpainting architecture.  The hole is created by removing the red bounding box region in the original image. Given a guidance image, our approach first localizes a matching patch, and then synthesizes a consistent hole-filling informed by the guidance patch. }
\label{fig:multiple2}
\end{figure}

We mention in Section \ref{section5.3} that we will show  in the appendix diverse inpainting results for a hole when given different guidance images. In Figure \ref{fig:multiple1} and Figure \ref{fig:multiple2}, we qualitatively show diverse inpainting results for both indoor and outdoor scenes using our localization and synthesis networks. 
Compared to methods based on hallucination from context \cite{efros1999texture,ctxEnc2016,yang2016high,iizuka2017globally,yu2018generative}, our approach can obtain diverse realistic hole-fills for a hole given different guidance images.

\subsection{Localization Failures}
\begin{figure}
\centering
\includegraphics[width=1.0\textwidth]{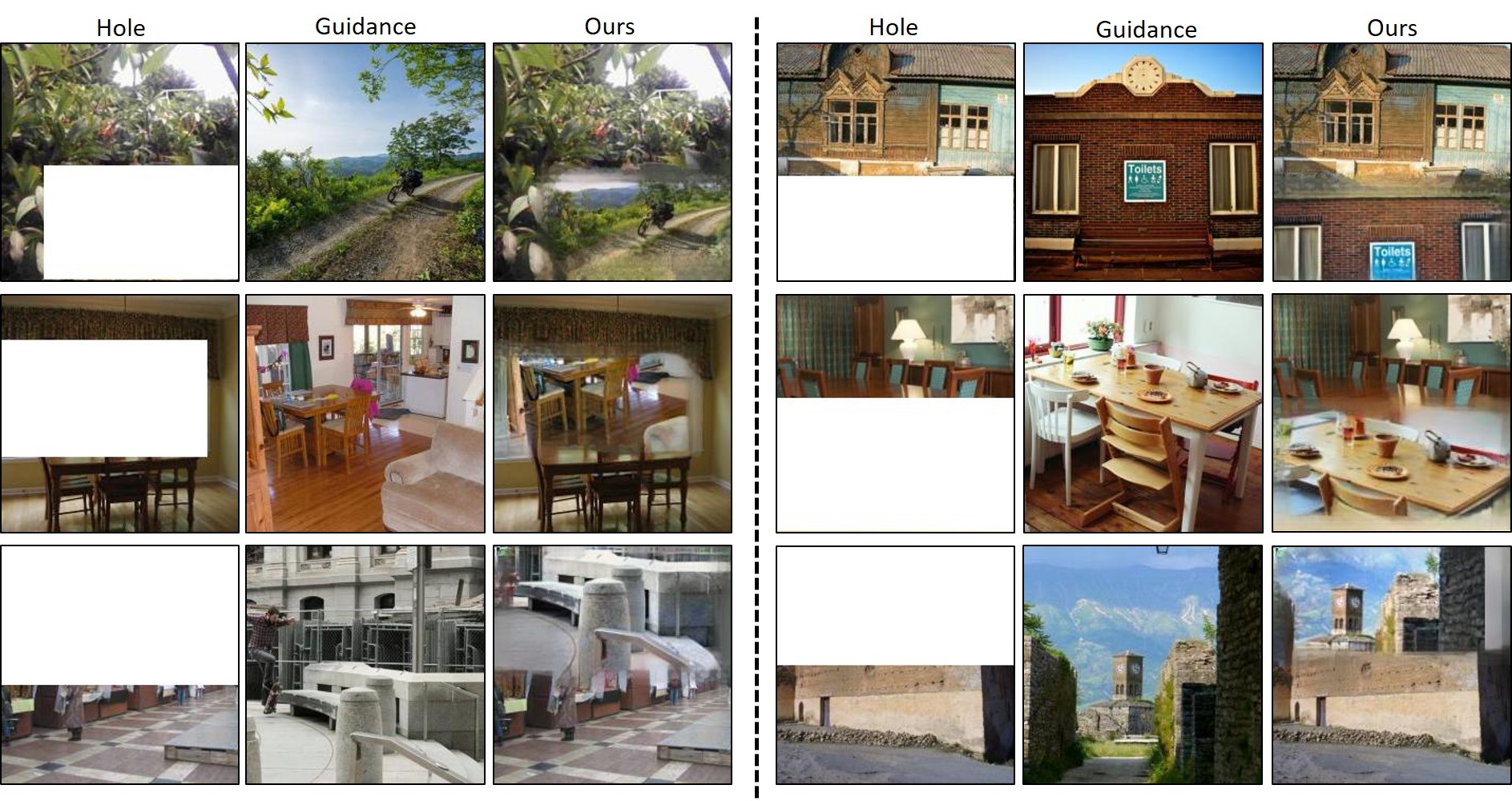}
\caption{We show failure cases of our localization network. When the localization network fails to localize a matching patch, the synthesis network cannot synthesize a reasonable hole-fill.}
\label{fig:limitation2}
\end{figure}

In Section \ref{section5.4}, we show that our localization is slightly better than \textit{Local Context Matching}~\cite{hays2007scene}. When our localization network fails to localize a matching patch, the synthesis network cannot synthesize a reasonable hole-fill, as exemplified in Figure~\ref{fig:limitation2}. Our relatively small gain over the baseline is probably because the distribution of features for localization in the synthetic images is different from that in natural images. The localization network trained on the synthetic dataset does not generalize to natural images as well as the synthesis network. 

\subsection{AMT}
In this section, we provide a more detailed description
of two types of AMT experiments we perform: limited-time absolute realism evaluation and unlimited-time relative realism comparison.
\subsubsection{Absolute Realism}
\begin{figure}
\centering
\includegraphics[width=0.97\textwidth]{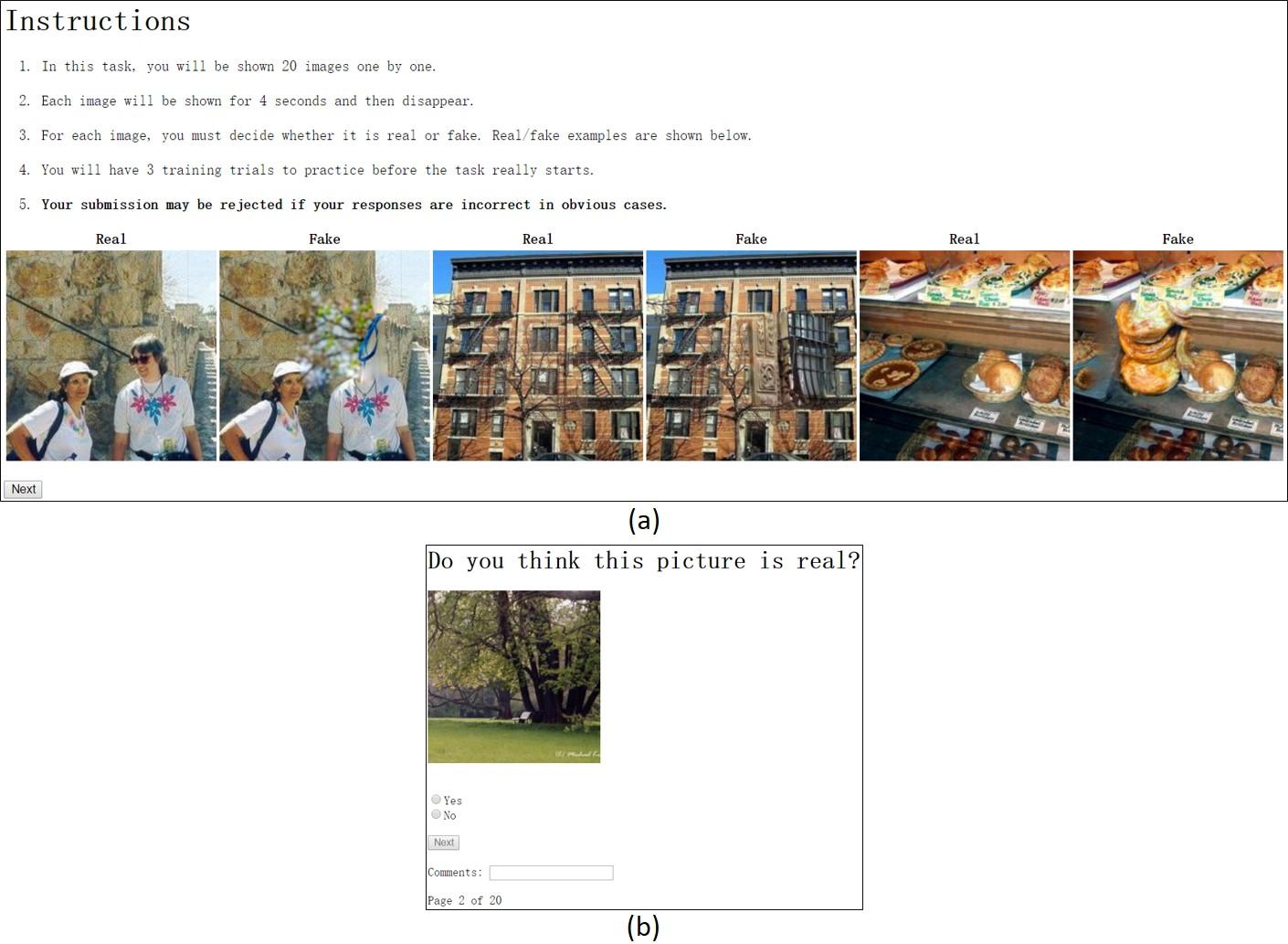}
\caption{We show the (a) instruction page and (b) interface of our AMT experiment on absolute realism.}
\label{fig:AMT_abs}
\end{figure}

We evaluate absolute realism of an image by asking people if the shown image is real or fake. Each HIT contains three training examples followed by 20 test images. Motivated in part by the work of Joubert et al. \cite{joubert2007processing}, we show each image for four seconds to give people sufficient time to determine its realism. Each HIT is completed by three unique workers. If two or more workers agree that the image shown is real, we label that image as real. Screenshots of the instruction page and interface are shown in Figure \ref{fig:AMT_abs}.

\subsubsection{Relative Realism}
\begin{figure}
\centering
\includegraphics[width=0.97\textwidth]{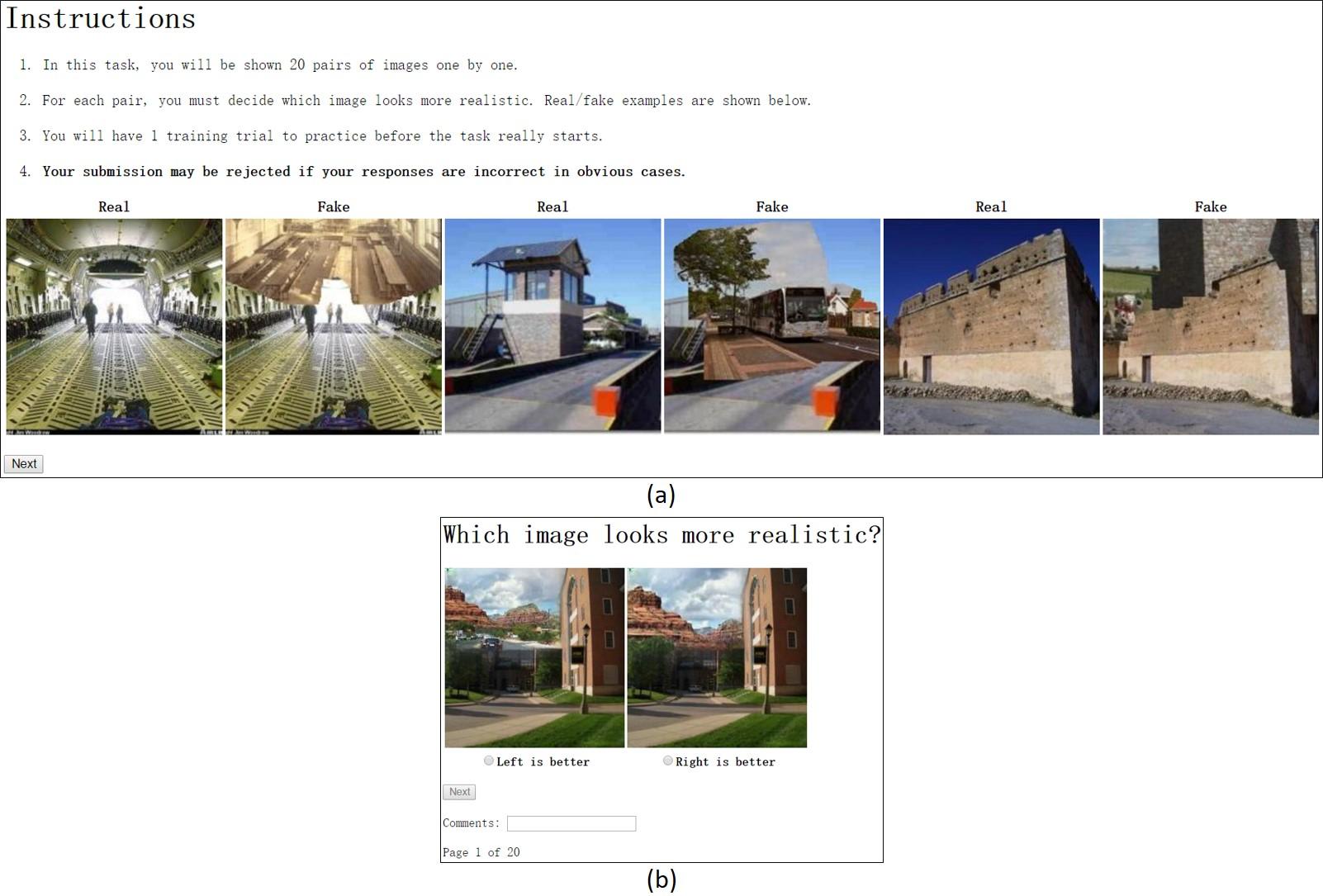}
\caption{We show the (a) instruction page and (b) interface of our AMT experiment on relative realism.}
\label{fig:AMT_relative}
\end{figure}

We draw a direct relative comparison between a pair of images by asking people to choose the more realistic one in two images shown side-by-side. Each HIT contains one training example followed by 20 test comparisons. Each image is shown for unlimited time to allow people to carefully examine the difference. Each HIT is completed by three unique workers. If two or more workers agree that image A is more realistic than image B, we label A as more realistic. Screenshots of the instruction page and interface are shown in Figure \ref{fig:AMT_relative}.

\end{document}